\documentclass[nohyperref]{article}

\usepackage{microtype}
\usepackage{graphicx}
\usepackage{subfigure}
\usepackage{booktabs} 

\usepackage{hyperref}

\usepackage[accepted]{icml2022}

\usepackage{amsmath}
\usepackage{amssymb}
\usepackage{mathtools}
\usepackage{amsthm}

\usepackage[capitalize,noabbrev]{cleveref}

\theoremstyle{plain}
\newtheorem{theorem}{Theorem}[section]

\theoremstyle{definition}

\theoremstyle{remark}

\usepackage[textsize=tiny]{todonotes}

\newcommand{\R}{\mathbb{R}}

\renewcommand{\[}{\begin{eqnarray}}
\renewcommand{\]}{\end{eqnarray}}

\icmltitlerunning{Dual Decomposition of Convex Optimization Layers for Consistent Attention in Medical Images}

\begin{document}

\twocolumn[
\icmltitle{Dual Decomposition of Convex Optimization Layers for \\
            Consistent Attention in Medical Images}



\icmlsetsymbol{equal}{*}

\begin{icmlauthorlist}
\icmlauthor{Tom Ron}{ie_tech}
\icmlauthor{Michal Weiler-Sagie}{rambam,med_tech}
\icmlauthor{Tamir Hazan}{ie_tech}
\end{icmlauthorlist}

\icmlaffiliation{ie_tech}{The Faculty of Industrial Engineering and Management, Technion Institute of Technology, Haifa, Israel}
\icmlaffiliation{rambam}{Department of Nuclear Medicine, Rambam Health Care Campus, Haifa, Israel}
\icmlaffiliation{med_tech}{Rappaport Faculty of Medicine, Technion Institute of Technology, Haifa, Israel}

\icmlcorrespondingauthor{Tom Ron}{tomron27@campus.technion.ac.il}
\icmlcorrespondingauthor{Tamir Hazan}{tamir.hazan@technion.ac.il}
\icmlkeywords{Machine Learning, ICML}

\vskip 0.3in
]



\printAffiliationsAndNotice{}  

\begin{abstract}
A key concern in integrating machine learning models in medicine is the ability to interpret their reasoning. 
Popular explainability methods have demonstrated satisfactory results in natural image recognition, yet in medical image analysis, many of these approaches provide partial and noisy explanations. Recently, attention mechanisms have shown compelling results both in their predictive performance and in their interpretable qualities. A fundamental trait of attention is that it leverages salient parts of the input which contribute to the model's prediction. To this end, our work focuses on the explanatory value of attention weight distributions. We propose a multi-layer attention mechanism that enforces consistent interpretations between attended convolutional layers using convex optimization. We apply duality to decompose the consistency constraints between the layers by reparameterizing their attention probability distributions. We further suggest learning the dual witness by optimizing with respect to our objective; thus, our implementation uses standard back-propagation, hence it is highly efficient. While preserving predictive performance, our proposed method leverages weakly annotated medical imaging data and provides complete and faithful explanations to the model's prediction. 
\end{abstract}

\section{Introduction}
\label{introduction}
Since the emergence of artificial intelligence, there has been great interest in machine learning algorithms' reasoning processes. In high-stakes domains, such as medical image analysis, understanding the inner workings behind a model's decision is crucial for building trust and accountability between end-users and algorithms. The field of explainable artificial intelligence (XAI) generally addresses two notions regarding model interpretation: the \textit{"how"} and the \textit{"why"}. The former is often focused on the model's architecture and optimization process, which grew to be challenging as models became complex. By contrast, the latter can be viewed from a data-centered perspective: which particular attributes of the input inclined the model towards a specific decision.


\begin{figure}[t]
    \centering
    \includegraphics[width=0.4\columnwidth]{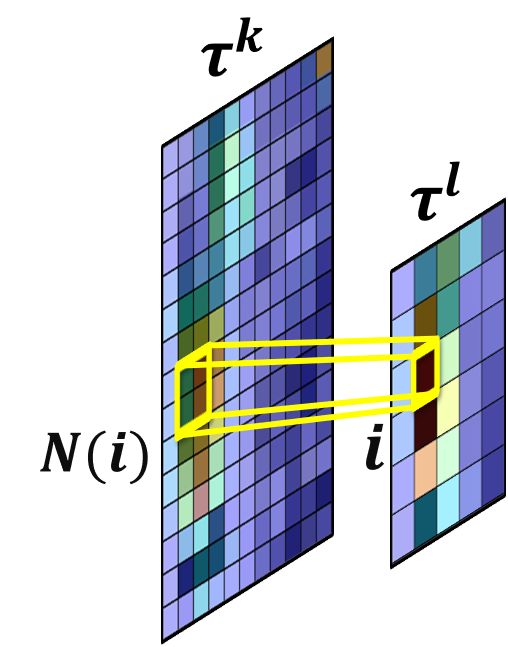}
    \caption{
    \small 
    We propose a multi-layer visual attention mechanism for obtaining comprehensive explanations in medical image recognition. One layer is closer to the input and it obtains a detailed attention of the input, while the other is closer to the output and  it obtains a detailed attention of the prediction. The decision is performed only on attended areas in the image and consequently we encourage comprehensive explanation. Using a spatial marginalization layer, we enforce consistency between the two attention units to ensure a consistent explanation.  
    }
    \label{fig:taus}
\end{figure}

To this end, a magnitude of explainability methods has been proposed in recent years, including local linear approximations, gradient-based methods, post-hoc explanations and attention-based explanations: A theoretically solid class of explainers are based on local linear approximations \cite{lime, intgrad, lrp, deeplift}. These methods often require a reference point for computing an explanation for a specific input. However, finding an appropriate reference point for each input is unwieldy, particularly in neural networks with non-linear activations \cite{xairev1}. Gradient-based methods \cite{intgrad, fullgrad, smoothgrad, deeplift, gradcam} are another popular class of explainers, but they often suffer from volatility \cite{shatt} or output coarse explanations. In addition, recent analysis has shown that in computer vision, gradient-based explanations degrade into edge-detectors \cite{sanity} and exhibit other pathologies \cite{path}. The majority of methods are typically applied post-hoc \cite{lrp, deeplift, intgrad, smoothgrad, fullgrad, gradcam, lime, shap}. That is, given a trained model and an input example, they observe the propagation of input throughout the model and process this observation into an explanation.

Unfortunately, such explanation methods have inherent weaknesses upon application in medical image analysis. As opposed to most tasks in natural image recognition, detecting pathologies in medical images typically requires expertise and years of training for humans. Features such as color, texture, and shape in the medical image landscape are often intricate and subtle to discern. Volatility and incompleteness of post-hoc methods can easily exacerbate and produce noisier explanations compared to the natural image setting. In addition, as modern-day convolutional nets are optimized solely for predictive performance, analysis of post-hoc explanations typically reveals that the models focus on the top-most discriminative features of the input. While this reliance on a small subset of features might provide plausible explanations for natural images, medical image analysis imposes stricter requirements. Often the medical physician would demand a broader glimpse into the algorithm's reasoning process, i.e., to be presented with \textit{all} the discriminative features available in the input. Our experiments demonstrates that this information is not well tractable using post-hoc explanation methods. In pursuit of this notion, \citet{rudin2019stop} argued against using post-hoc explanations for black-box models (particularly in high-stakes settings) and advised using interpretable models instead. Our work is motivated by such criteria, where we use intrinsic attention to produce reliable and complete interpretations.


A prominent class of machine learning algorithms is attention-based models \cite{bahdanau2014neural, aiayn, bert}. Alongside high predictive performance, these models admit transparency through their attention weight distributions. Recent works have discussed whether attention distributions can be considered as explanations to model prediction \cite{attn_isnot, attn_isnotnot, fresh, yu}, resolved with the notions of plausible, faithful and comprehensive explanations. These works are focused on natural language processing while our work aims at learning comprehensive explanations using attention for visual models, i.e, explanations that contain all the relevant information for the model's prediction.  

In this work, we propose a multi-layer visual attention mechanism for obtaining comprehensive explanations in medical image recognition. To enforce comprehensive explanations we learn two attention layers: one layer is closer to the input and it obtains a detailed attention of the input, while the other is closer to the output and  it obtains a detailed attention of the prediction. The decision is performed only on attended areas in the image and consequently we encourage comprehensive explanation. Using a spatial marginalization layer, we enforce consistency between the two attention units to ensure a consistent explanation. Using dual decomposition, we demonstrate that our constraint admits an optimal solution that is equivalent to a solution obtained by a convex optimization program, yet it is achieved efficiently by end-to-end back-propagation. We assert the quality of explanations produced by our approach in two experiments, one conducted on a self-curated, novel dataset consisting of 2637 labeled images of $^{18}$F-Fluorodeoxyglucose Positron Emission Tomography/Computed Tomography (FDG-PET/CT) patient scans.

\section{Related work}



\textbf{Explanation by attribution propagation.} Attribution propagation methods follow an iterative scheme where a model's output (representing an initial relevance score) is back-propagated and decomposed across layers of the model. Formalized by \citet{dtd}, these methods apply a Taylor expansion at each layer of the network, and attribution is repeatedly distributed until reaching the input. Layer-Wise Propagation (LRP) \cite{lrp} backpropagates a prediction score while weighting contributions from each layer in networks using ReLU activations. DeepLIFT \cite{deeplift}  decomposes the output by computing differences of contribution scores between the activation of each neuron to some reference point. Attribution propagation methods are typically fast to compute and provide interpretation at the pixel level. 

Other attribution techniques include saliency and occlusion \cite{saliency_gal, saliency_mahendran, saliency_simonyan, saliency_zeiler, saliency_zhou}, architectural modifications \cite{excit} and input perturbation \cite{input_perturb, input_perturb2}. In particular, popular methods such as LIME \cite{lime}, Shapley-value sampling \cite{shap} and L2X \cite{l2x} offer solid theoretical frameworks for model interpretation, but are computationally heavy as they require multiple evaluations or substantial explanation layers.

\textbf{Explanation by gradient analysis.} Gradient-based methods leverage gradients computed during back-propagation to obtain a relevancy score for the input of each layer or neuron. The magnitude and sign of gradients can be visualized, thus providing an explanation for which parts of a layer's input were most influential to its output. Gradient$\times$Input \cite{deeplift} multiplies the input to each layer by the partial derivatives of its output. \citet{towards} demonstrated that under ReLU activations, a zero-baseline and no  biases, LRP \cite{lrp} and DeepLIFT \cite{deeplift} are equivalent to Gradient$\times$Input. Integrated Gradients \cite{intgrad} assign importance to input features by approximating the integral of gradients with respect to the inputs along a path from a given baseline. SmoothGrad \cite{smoothgrad} and FullGrad \cite{fullgrad} attempt to stabilize gradient-based explanations by averaging, adding Gaussian noise, and other aggregative manipulations. Notably, Class Activation Mapping (CAM) approaches such as GradCAM \cite{gradcam} combine the activation and gradients of a given layer, conditioned on a target class.  GradCAM provides compelling visual results but usually only applicable to very deep layers, resulting in coarse heatmaps. 


\textbf{Attention and explainability.} Since introduced by \citet{bahdanau2014neural}, attention mechanisms have grown ubiquitous in image recognition \cite{ltpa, self_attn1, self_attn2, self_attn3, self_attn4, self_attn5, self_attn6} and in multi-modal tasks such as visual question answering \cite{vqa2, vqa3, high_order} and visual dialog \cite{visdial, fga, aware}. \citet{oktay2018attention} demonstrated the predictive benefits of visual attention in medical image segmentation. In natural language processing, transformers \cite{aiayn, bert} rapidly became the backbone to many state-of-the-art models in machine translation \cite{trans_wang, trans_mehta}, question answering \cite{trans_luke} and sentiment analysis \cite{trans_jiang, trans_raffel, trans_lan}.
Promising attempts to adjust transformers toward tasks in computer vision have taken place. \citet{vit} introduced the Visual Transformer (ViT), which treats patches of the input image as tokens. Although being resource-intensive, it demonstrated competitive results with state-of-the-art convolutional networks in classifying natural images.

Inspired by bottom-up/top-down functions of cognitive processes \cite{connor2004visual}, attention mechanisms are often designed to focus on input features that are significant to the model's prediction. Considering whether attention weight distribution can rightfully explain the model's prediction was at the center of a recent discussion. \citet{attn_isnot} empirically demonstrated that drastically distorting attention weights has little effect on model performance, hence considering these to have explanatory value is spurious. In return, \citet{attn_isnotnot} refined the discussion and considered two types of explanations: plausible and faithful. Plausible explanations are described as such that would seem intuitive to a human being, e.g., a heatmap highlighting \textit{some} discriminative features of a target class. Faithful explanations guarantee that {\em only} the highlighted features were used in the model's prediction; therefore, they can be causally associated with the prediction. 
\citet{yu} has further extended these criteria into the notion of comprehensiveness: explanations must contain \textit{all} the relevant information for the model's prediction. Along with the demand of \citet{rudin2019stop} to devise models with internal explanations, we follow the faithfulness and comprehensiveness paradigms in the design of our method.

\textbf{Convex optimization.} Convex optimization methods have been recently integrated into deep neural networks. \citet{amos} introduced OptNet, a neural architecture that can encode quadratic programs as trainable layers. \citet{convex_layer} proposed a new grammar and translation form for convex problems, allowing to automatically differentiate through convex layers using standard back-propagation. \citet{domke}, \citet{sst} and  \citet{berthet} offered to compute gradients to non-differentiable operations by perturbation, which in turn can be used to back-propagate with respect to a solution of a convex problem. Nevertheless, differentiation by perturbation and convex optimization layers poses significant computational burdens. The former often involves multiple sampling schemes and generates biased gradients, and the latter is typically solved synchronously while heavily relying on sequential computations. Our work differs from the aforementioned works, as we use convex duality to decompose the optimization and rely on the dual reparameterization to ensure optimality. Moreover, we suggest learning the dual witness within the deep net; thus, we can enjoy the benefits of parallelized GPU operations. Additional overview on machine learning and dual decomposition can be found in \cite{rush2012tutorial, sra2012optimization}.


\begin{figure*}[t]
    \centering
    \includegraphics[width=14cm]{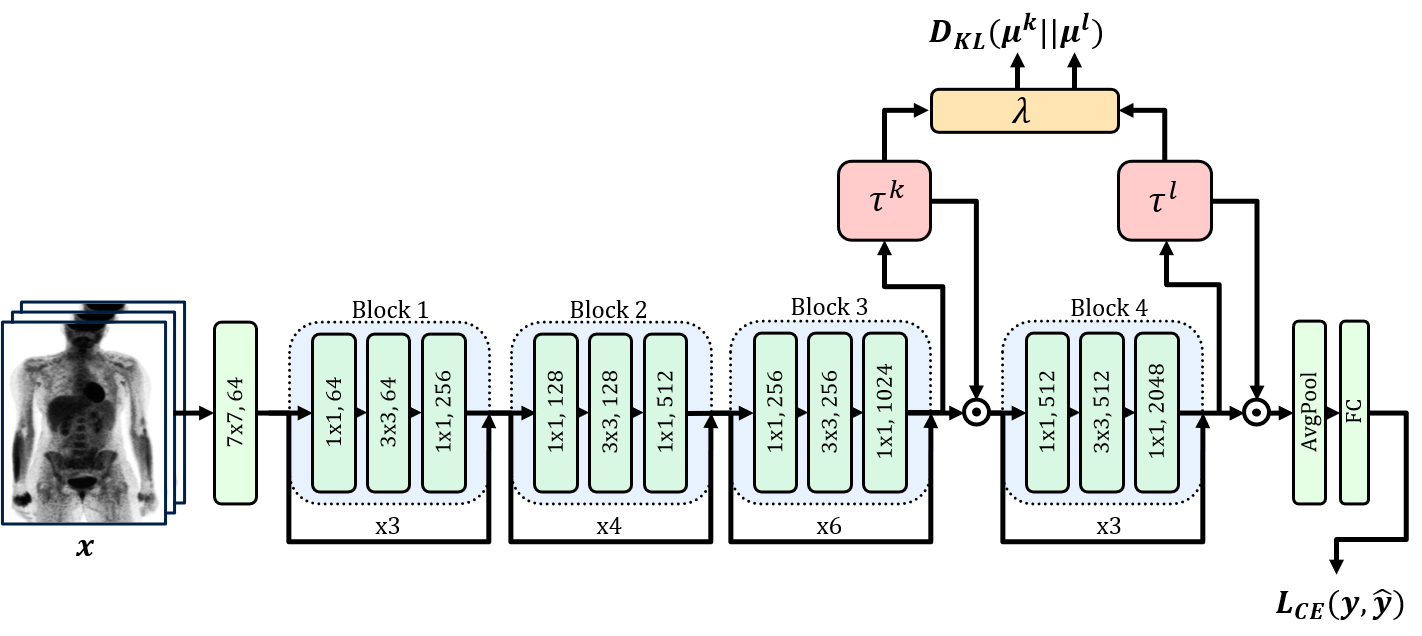}
    \caption{\small ResNet-50 architecture with attention gates ($\tau^k, \tau^l$) and a marginal consistency layer $\lambda$. Enforcing the marginalization consistency using convex optimization layer (see Eq.~\ref{eq:cvx}) couples the whole block into a single optimization layer. The duality theorem allows to decompose these layers and efficiently enforce this constraint using reparameterization (see Theorem~\ref{marginals_theorem}).}
    \label{fig:arch}
\end{figure*}

\section{Explaining decisions on medical images using consistent attention} \label{method}

We learn consistent attention probability distributions to explain machine learning decisions in medical image recognition. Our primary task is learning to classify levels of altered biodistribution of FDG in PET/CT scans while providing comprehensive interpretations to the algorithm's prediction. To encourage comprehensive explanations we restrict the model to rely only on attended neurons while preserving the spatial structure of the image. We use the attention probability model to weigh the amount of visual information that flows through the convolutional layers.

A convolutional layer is a 3-dimensional objects of dimension $n \times n \times d$. The $n \times n$ cells corresponds to $n \times n$ sub-windows in the image. The spatial location of each sub-window corresponds to a convolutional neuron and is indexed by $i = 1,...,n^2$. The $i^{th}$ convolutional neuron is described by the vector $v_i \in \R^{d}$ which represents the corresponding sub-window in the image. In the following, we refer to the $k^{th}$ convolutional layer by an additional index, i.e., the dimension of the $k^{th}$ layer is $n_k \times n_k \times d_k$ and the $i^{th}$ sub-window representation is denoted by $v_i^k \in \R^{d_k}$. 

We learn an attention probability distribution over intermediate convolutional layers to weigh the information of their $i^{th}$ neuron. Consequently, we are able to emphasize the information in the relevant sub-windows using the learned attention probability distribution. Formally, for the $k^{th}$ convolutional layer, we learn a probability distribution $\tau^k_i$ over the $n_k \times n_k$ sub-windows of the input image. Given $v^k_i \in \R^{d_k}$, we learn $\tau^k_i$ by first learning its potential function 
\[
\phi^k_i = \left \langle {u^k_i}, \mbox{relu}\left( \frac{U_k v^k_i}{||U_k v^k_i||} \right) \right \rangle
\] \label{eq:potential}
where $U_k \in \mathbb{R}^{d_k \times d_k}$ and $u^k_i \in \mathbb{R}^{d_k}$ are learnable parameters. 

The attention probability distribution is set to be its potential function softmax, namely, 
\[
\tau^k_i = \frac{e^{\phi^k_i}}{\sum_{j=1}^{n_k^2} e^{\phi^k_{j}}}.
\] \label{eq:tau} 
Intuitively, this attention probability distribution assigns high probabilities to sub-windows that are important in the learner's decision and assigns low probabilities to sub-windows that do not contribute to the decision. By doing so, visualizing the attention probability distribution introduces spatial information to the medical physician, accounting for areas in the image that the learner relies on in its decision making (see Fig. \ref{fig:taus}). 

Our spatially attended embedding, $\hat v^k_i$, of the $k^{th}$ layer combines both the semantic information in the image as well as the spatial information about the location of the attention distribution. The semantic information is represented by the vectors $v^k_i \in \R^{d_k}$ while the spatial information is represented by the attention probability distribution $\tau^k_i$. We combine these two components using the pointwise multiplication: $\hat v^k = \tau^k \odot v^k$, which is defined by 
\[
\hat v^k_i = \tau^k_i \cdot v^k_i.
\] \label{eq:attended}
The dimension of the attended embedding $\hat v^k_i$ is identical to the dimension of the image embedding $v^k_i$. Intuitively, the attention probability distribution $\tau^k_i$ has high values for relevant $i^{th}$ sub-windows of the image, i.e., sub-windows that contain semantic information for the medical diagnosis task. Equivalently $\tau^k_i \approx 0$ for irrelevant sub-windows. Hence, the attended embedding in Eq. (\ref{eq:attended}) consists of the vector $\hat v^k_i \approx 0$ whenever $\tau^k_i \approx 0$, i.e., for sub-windows that are irrelevant for the model's decision in the $k^{th}$ layer. Equivalently, for semantically meaningful sub-windows, the attended embedding is similar to the original image embedding, i.e., $\hat v^k_i \approx v^k_i$. This embedding highlights the spatial locations of the semantically meaningful sub-windows and populates them with the respective semantic information of the image. In addition, the embedding restricts the decision of the deep net to rely on sub-windows for which their corresponding attention probability distribution is high. The attention probability distribution, in turn, provides meaningful spatial information to the medical physician interpreting the scan and points towards the area that the decision relied on. Thus, it serves as a faithful visual explanation for the algorithm's decision in the image.

\subsection{Consistent Attention} 

To enforce comprehensive explanations we learn two attention layers in a standard convolutional nets (e.g., ResNet-50 \cite{resnet}). Due to the convolutional net structure, these attention probability distributions play different roles in explaining network prediction. The first layer, $\tau^k$, is closer to the input therefore it better represents variability in the data, e.g., gradients in the image. Also, a convolutional layer that is closer to the input has more convolutional neurons and therefore has a higher resolution with respect to the input image. The second layer, $\tau^l$ for $k < l$, is closer to the output and provides semantic information about the predicted label.

We suggest to use convex optimization to obtain consistent attention across $\tau^k, \tau^l$. We use the spatial relations between layers to enforce marginalization consistency constraints between $\tau^k$ and $\tau^l$, as illustrated in Fig. (\ref{fig:arch}). We define $N(i)$ to be the set of indexes $j \in N(i)$ for which $\tau^k_j$ corresponds to $\tau^l_i$. In Fig. (\ref{fig:taus}) we can see that $j \in N(i)$ whenever $j$ and $i$ refer to the same sub-window in the image. To guarantee consistent attention, we would like to enforce the marginalization constraint $\sum_{j \in N(i)} \tau^k_j = \tau^l_i$ for every $i$. To do so, we infer consistent probability distributions $\mu^k, \mu^l$ that are closest to $\tau^k, \tau^l$ respectively and satisfy the marginalization constraints. We use the $KL$-divergence $D_{KL}(\mu || \tau) = \sum_i \mu_i \log(\mu_i / \tau_i)$ to measure the similarity between $\mu$ and $\tau$. Thus, the consistent attention probability model is the output of the following convex  program:

\vspace{-0.5cm} 
\begin{equation}
  \label{eq:cvx}
  \begin{gathered}[b]
    \arg \min_{\mu^k,\mu^l} D_{KL}(\mu^k || \tau^k) + D_{KL}(\mu^l || \tau^l) \hspace{0.5cm} \\
    \mbox{s.t.}  \hspace{0.2cm} \mu^k, \mu^l \hspace{0.2cm} \mbox{are distributions, and} \hspace{0.2cm} \mu^l_i = \sum_{j \in N(i)} \mu^k_j
  \end{gathered}
\end{equation}
\vspace{-0.5cm} 

Unfortunately, it is computationally challenging to solve this program as a convex optimization layer as it couples all parameters from layer $k$ to layer $l$ by the same convex program (see Fig. \ref{fig:arch}). 

\subsection{Dual decomposition and primal-dual optimality conditions} 
\label{dualdecomp}
We propose to use strong duality to decompose the optimization that couples the $k^{th}$ layer and the $l^{th}$ layer. Lagrange duality theorem relies on Lagrange multipliers $\lambda$ that reparameterize the primal variables $\tau^k$ and $\tau^l$ in order to obtain the optimal solutions $\mu^k, \mu^l$ of the primal program in Eq. (\ref{eq:cvx}). Our reparameterization is attained by relating a Lagrange multiplier $\lambda_i$ to each marginalization constraint $\mu^l_i = \sum_{j \in N(i)} \mu^k_j$. We use the enhanced Fritz-John condition \cite{fritz} to enforce the probability distribution constraints for $\mu^k, \mu^l$ implicitly. Therefore, we are able to attain the following reparameterization for the optimal solution of Eq. (\ref{eq:cvx}): 
\begin{theorem} \label{marginals_theorem}
The probability distributions $\mu^k, \mu^l$ are the optimal solutions to the primal program in Eq. (\ref{eq:cvx}) if 
(i) $\mu^l_i = \frac{\tau^l_i e^{\lambda_i}}{\sum_s {\tau^l_s e^{\lambda_s}}}$; (ii)  $\mu^k_j = \frac{\tau^k_j e^{-\lambda_i}}{\sum_t \sum_{s \in N(t)} \tau^k_s e^{-\lambda_t}}$; and (iii) $\mu^l_i = \sum_{j \in N(i)} \mu^k_j$, namely, the parameterized versions $\mu^l, \mu^k$ agree on their marginal distributions.  
\end{theorem}
\begin{proof}
See Section 3.2 of the appendix.
\end{proof}
Strong duality allows us to decompose the constraint across the $k^{th}$ and the $l^{th}$ layers using reparameterization. That is, whenever we are able to find a $\lambda$ for which the two distributions $\frac{\tau^l_i e^{\lambda_i}}{\sum_s {\tau^l_s e^{\lambda_s}}}$ and $\frac{\tau^k_j e^{-\lambda_i}}{\sum_t \sum_{s \in N(t)} \tau^k_s e^{-\lambda_t}}$ agree on their marginals, then these distributions are guaranteed to be primal optimal and to propagate along the deep net. Summarizing these conditions, we  search for the following reparameterized condition 
\[
\sum_{j \in N(i)} \frac{\tau^k_j e^{-\lambda_i}}{\sum_t \sum_{s \in N(t)} \tau^k_s e^{-\lambda_t}} = \frac{\tau^l_i e^{\lambda_i}}{\sum_s \tau^l_s e^{\lambda_s}}.  \label{eq:reparameterization}
\]
The above reparameterization use the dual variables $\lambda$ to guarantee the primal-dual optimality conditions in Theorem \ref{marginals_theorem}. Hence $\lambda$ serves as a dual witness to the optimality of the primal variables $\mu^l, \mu^k$ in Eq.~(\ref{eq:cvx}) if their reparameterization satisfy their marginalization constraints.   

\subsection{Learning the dual witness}
Unfortunately, the reparameterized constraint in Eq. (\ref{eq:reparameterization}) needs to hold for any data point in training and in particular in any batch operation. While this can be done by projected stochastic gradient descent, this is computationally unappealing. Instead, we suggest learning $\lambda$ for every pair of $\tau^k, \tau^l$. We do so by introducing a shallow layer, parameterized by $\lambda$, which encodes the two counterparts of Eq. (\ref{eq:reparameterization}).
We then enforce the constraint by adding a penalty function to our optimization objective

\begin{equation}
\label{eq:loss}
    \mathcal{L}_{CE}(y, \hat{y}) +  
    \alpha D \Big(\frac{\tau^l_i e^{\lambda_i}}{\sum_s \tau^l_s e^{\lambda_s}} , \sum_{j \in N(i)} \frac{\tau^k_j e^{-\lambda_i}}{\sum_t \sum_{s \in N(t)} \tau^k_s e^{-\lambda_t}} \Big)
\end{equation}

Where $\mathcal{L}_{CE}(\cdot, \cdot)$ is the categorical cross-entropy loss and $D(\cdot, \cdot)$ is a divergence function, e.g., $KL$-divergence, which encourages the reparameterization condition in Eq.~(\ref{eq:reparameterization}). The positive number $\alpha > 0$ controls the loss incurred when the reparameterization condition is not satisfied. 



\begin{figure}[t]
    \centering
    \includegraphics[width=\columnwidth]{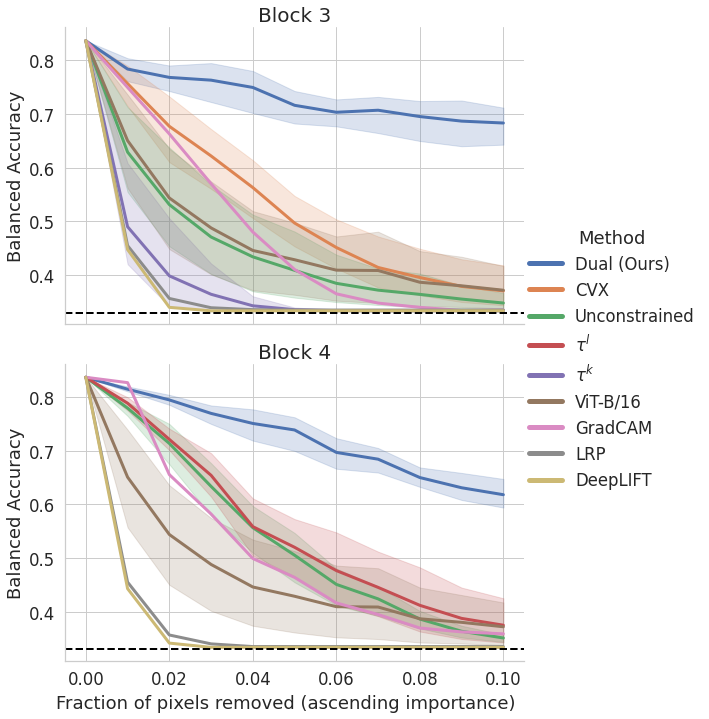}
    \caption{\small Input perturbation curves on FDG-PET-MUAB dataset. 
    Conceivably, faithful explanations only highlight relevant regions in the input image; thus, removing irrelevant regions should have a lesser impact on downstream accuracy. Compared to the baselines, the classifier demonstrates a significantly lesser decay in accuracy when removing irrelevant pixels according to our \textit{Dual} approach.}
    \label{fig:perturb_lineplot}
\end{figure}

\section{Experiments} \label{experiments}
We evaluate the explanatory value of the attention distribution obtained by our method in two experiments on two medical imaging datasets. 
In each experiment, we train a vanilla CNN model and extract attribution maps using various baselines. In addition, we compare our visual explanations to attention weights obtained from a fine-tuned visual transformer (ViT-B/16) which was pretrained over ImageNet \cite{rw2019timm}. We divide the baselines into three subgroups: gradient-based (GradCAM \cite{gradcam}; Gradient$\times$Inputs \cite{deeplift}), attribution propagation-based (DeepLIFT \cite{deeplift}; LRP \cite{lrp}) and attention-based (ViT-B/16 \cite{vit}; single-gate and unconstrained attention (Eq. \ref{eq:potential}); CVX attention (Eq. \ref{eq:cvx}); dual decomposition of consistent attention (Eq. \ref{eq:loss})). For methods that are class dependant \cite{deeplift, gradcam}, we condition the attribution on the ground truth label. For methods that provide channel-level attribution \cite{deeplift, lrp}, we perform 0-1 normalization on each channel and sum attributions across channels. We compare attention attribution maps from several convolutional layers to the class activation maps of the equivalent layers in \cite{gradcam}, to sum-pooled attribution maps in \cite{deeplift, lrp} and to an average of attention heads weights in \cite{vit}. Our models and baselines were implemented using PyTorch \cite{pytorch} and Captum \cite{kokhlikyan2020captum} and were trained on a single Nvidia GeForce RTX 2080TI GPU.

\begin{table}
\begin{center}
\caption{\small Area under the curve (AUC) for input perturbation results, FDG-PET-MUAB dataset. We compute the area under the curves depicted in Fig. (\ref{fig:perturb_lineplot}) for a full range of fractions (0.0-1.0). Our \textit{Dual} approach drops to random accuracy when roughly 60\% of the input pixels are removed (by order of importance).} \hspace{\linewidth}

\label{tab:perturb_auc}


\resizebox{0.6\columnwidth}{!}{
\begin{tabular}{l|cc}
\toprule
Method          &       Block 3     &       Block 4     \\
\midrule
DeepLIFT        &   20.37   &   20.37   \\
LRP             &   20.48   &   20.48   \\
GradCAM         &   21.52   &   21.90   \\
\midrule
ViT-B/16        &   21.63$\pm$1.17   &   21.63$\pm$1.17   \\
$\tau^k$        &   20.52$\pm$0.34   &   -   \\
$\tau^l$        &   -   &   22.34$\pm$0.70   \\
Unconst.        &   21.23$\pm$0.72   &   22.11$\pm$0.37   \\
CVX             &   22.18$\pm$0.79   &   -   \\
Dual (Ours)     &   \textbf{32.48$\pm$1.81}   &   \textbf{27.59$\pm$1.23}   \\
\bottomrule
\end{tabular}
}
\end{center}
\vspace{-0.5cm}
\end{table}

\subsection{Input perturbation} \label{pet_exp}
Our first task is to measure the explanation that is provided by our consistent attention compared to explanation baselines when learning to classify altered biodistribution levels in FDG-PET/CT patient scans. The biodistribution is categorically labeled to normal, mild, and severe and depicts the diffusion of FDG radiotracer, which serves as a marker for the tissue uptake of glucose, which in turn is closely correlated with certain types of tissue metabolism. The explanation in this experiment directs the interpreting physician to areas in the image with normal, mildly, and severely altered muscular FDG uptake.

\textbf{Dataset:} The $^{18}$F-Fluorodeoxyglucose Positron Emission Tomography Muscular Uptake Altered Biodistribution (FDG-PET-MUAB) dataset contains 2637 images of 2470 patients that underwent a FDG-PET/CT scan in a nuclear medicine institute of a central health care campus during 2019 for various indications. Every image is a processed 2D coronal view of a maximum intensity projection (MIP) produced from a 3D PET DICOM file. For additional information regarding data collection, anonymization and annotation, please refer to the appendix.

\textbf{Experimental setup:} A baseline ResNet-50 \cite{resnet} model was trained on a randomly sampled train set (80/20, patient disjoint) and obtained a balanced accuracy score of 83.6\% on the test set. We used the balanced accuracy score as mildly and severely altered biodstiribution cases are typically infrequent. Next, we fine-tuned four variants of our method: $\tau^k$ with a single attention unit at the end of block 3; $\tau^l$ with a single attention unit at the end of block 4; \textit{Unconstrained} with two attention units ($\tau^k$, $\tau^l$) without a consistency term; \textit{Dual} with two attention units and a consistency loss term  (see Fig. \ref{fig:arch}). 

Each attention unit is comprised of two cascaded potential gates (Eq. (\ref{eq:potential})). The attention units output two spatial distributions ($\tau^k \in \mathbb{R}^{32\times32}, \tau^l \in \mathbb{R}^{16\times16}$) highlighting salient regions in the input image. In addition, using $\tau^k, \tau^l$ from the unconstrained variant, we solve Eq. (\ref{eq:cvx}) using a convex solver \cite{pyomo} for each sample in the test set and output $\mu^k, \mu^l$ as attribution maps (referred to as CVX model). The dual decomposition model was trained using the loss from Eq. (\ref{eq:loss}), setting $\alpha=10$ to match the average magnitude of loss terms. The rest of the models were trained using categorical cross-entropy loss. Adam \cite{kingma2014adam} optimizer was used, with a learning rate of $10^{-3}$ for all models. ViT-B/16 was fine-tuned using default hyper-parameters (12 attention heads and no dropout). Gradient and attribution propagation methods were computed with respect to the baseline model. The results of all fine-tuned models were averaged for five runs.

Following the evaluation scheme in \cite{negperturb, path, chefer2020transformer}, given an input $x$ and an attribution map, we rank the map elements by ascending importance. Subsequently, we perturb the input $x$ by removing a growing amount of pixels according to the ranked attribution. We compare our method to the baselines by feeding the perturbed input back into the baseline ResNet-50 model and measuring the balanced accuracy score.

\begin{figure*}[t]
    \centering
    \includegraphics[width=\textwidth]{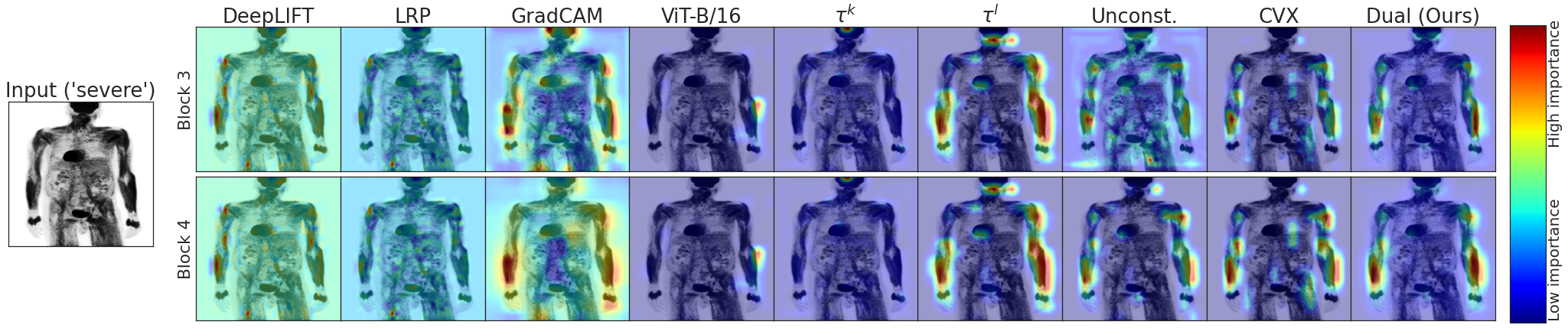}
    \caption{\small Explanations on a FDG-PET-MUAB dataset sample. The input image (leftmost) is labeled with severely altered biodistribution, which is evident mainly due to diffused uptake of FDG in both forearms, biceps, and deltoids. Compared to the baselines, we illustrate that our \textit{Dual} approach (rightmost) assigns both high importance (in red) to target regions and near-zero importance (in blue) to other regions, thus providing a comprehensive and faithful explanation. Visualizations of single-level attribution maps ($\tau^k$, $\tau^l$) are duplicated across the opposite block.}
    \label{fig:pet_sample}
    \vspace{-0.3cm}
\end{figure*}

\textbf{Results:} Fig. (\ref{fig:perturb_lineplot}) shows the input perturbation curve of our proposed \textit{Dual} approach compared to explanations from attention variants and additional baselines. Tab. (\ref{tab:perturb_auc}) shows the area under the curve for extended input perturbation curves from Fig. (\ref{fig:perturb_lineplot}). Fig. (\ref{fig:pet_sample}) illustrates visual explanations on a FDG-PET-MUAB dataset sample, obtained by our method and the baselines. Additional illustration of visual explanations can be found in the appendix.

\subsection{Implicit segmentation} \label{mri_exp}
Our second task is to measure the explanation obtained by our consistent attention compared to explanation baselines in magnetic resonance imaging (MRI). Given an axial 2D MRI slice of a brain hemisphere (left or right), the model learns to predict whether the hemisphere has a tumor or not. The explanation in this experiment directs the interpreting physician to the tumor's spatial location.

\textbf{Dataset:} The Brain Tumor Segmentation Challenge (BraTS18) dataset \cite{brats1, brats2, brats3} is a public dataset containing 285 MRI scans depicting pre-operative low and high-grade glioblastoma (LGG/HGG). Every sample includes four image modalities and a manually annotated segmentation mask. To obtain non-localized classification labels, we perform a vertical crop on every axial MRI slice, separating the tumor from the rest of the brain. As glioblastomas are usually common in one hemisphere, this produced two images from a given slice, each roughly with one hemisphere and a "yes" or "no" tumor label. We omitted the segmentation masks during training and included only binary labels as supervision. We discarded slices with low tumor area (less than $1000$ pixels) and performed a random 80/20 train-test split (patient disjoint).

\textbf{Experimental setup:} A baseline CNN using a U-Net encoder \cite{unet} was trained in a binary classification setting and achieved an accuracy score of 97.17\% on the test set. Similarly to Section \ref{pet_exp}, we fine-tuned four variants of our attention units at two mid and high-level convolutional blocks of the network, outputting up to two spatial distributions ($\tau^k \in \mathbb{R}^{64\times64}$, $\tau^l \in \mathbb{R}^{32\times32}$). In addition to Section \ref{pet_exp}, we trained two variants in which we replaced the second loss term in Eq. \ref{eq:loss} with a non-decomposed $\text{KL}$ term. This term directly enforces consistency between the two attention blocks and is equivalent to setting $\lambda = 0$. Baselines and hyper-parameters are identical to the ones described in Section \ref{pet_exp}. We compare our method to the baselines by measuring the overlap between the scaled attribution and segmentation mask annotations. As attribution maps output "soft" boundaries to areas of interest, we threshold the attribution maps by three quantiles ($0.975, 0.95, 0.9$) and measure the intersection over union (IoU) between the thresholded attribution and the segmentation mask. Furthermore, we compute the mean average precision score (mAP) between the attribution and segmentation masks, which is a threshold-agnostic metric. Our code for this experiment is publicly available\footnote{\url{https://github.com/tomron27/dd_med}}.

\begin{figure*}[t]
    \centering
    \includegraphics[width=\textwidth]{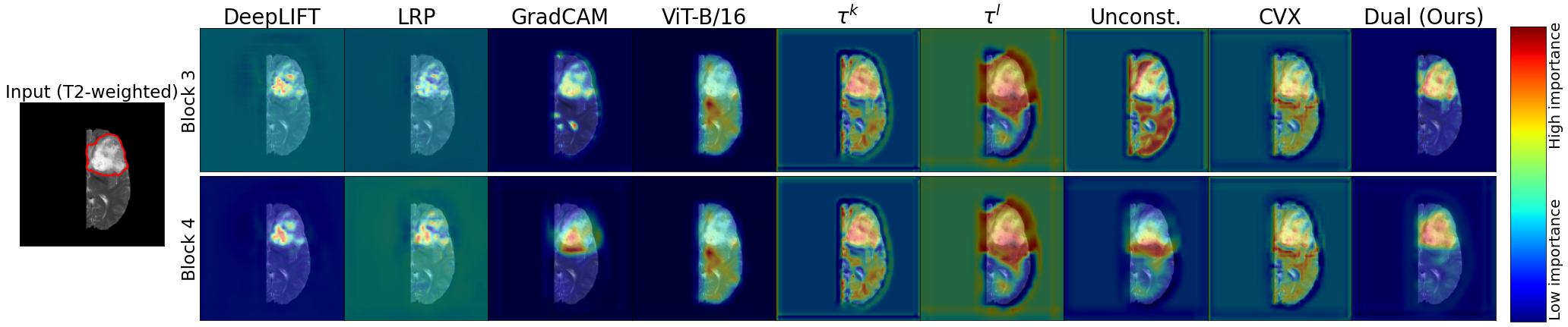}
    \caption{\small Brain tumor explanations on a vertically cropped BraTS18 dataset sample (positive instance). Input image and ground truth segmentation is in red contour (leftmost). Our \textit{Dual} approach (rightmost) obtains the highest overlap with the ground-truth segmentation mask (which was not provided as supervision). Other attribution methods produce partial or noisy explanations. Visualizations of single-level attribution maps ($\tau^k$, $\tau^l$) are duplicated across the opposite block.}
    \label{fig:mri_sample}
\end{figure*}

\textbf{Results:} Tab. (\ref{mri_table}) summarizes the mAP and IoU metrics for explanations obtained from our proposed \textit{Dual} approach compared to explanations from attention variants and additional baselines. Fig. (\ref{fig:mri_sample}) illustrates visual explanations on a BraTS18 dataset sample obtained by our method and the baselines. An additional illustration of visual explanations can be found in the appendix.



\begin{table}[ht!]
\begin{center}
\caption{\small mAP and IoU results for implicit segmentation on the BraTS18 MRI dataset. We collect visual explanations (attributions) from our \textit{Dual} method and multiple baseline explainers in two convolutional blocks (3,4). We measure the overlap between the explanation and ground truth segmentation at different spatial resolutions. Overlap was computed against positive halves of the source image (which contain a tumor). We compare our \textit{Dual} approach against the baselines.} \hspace{\linewidth}
\label{mri_table}
\resizebox{\columnwidth}{!}{
\setlength{\tabcolsep}{2pt}
\begin{tabular}{l|cccc}
\toprule
                 & \multicolumn{4}{|c}{Block 3} \\
\midrule
Method          & $mAP$ & $IoU_{0.975}$ & $IoU_{0.95}$ & $IoU_{0.90}$ \\
\midrule
DeepLIFT        & 49.40 & 35.18 & 29.20 & 18.78 \\
LRP             & 35.61 & 25.12 & 22.03 & 15.18 \\
GradCAM         & 58.68 & 31.59 & 39.34 & 34.20 \\
\midrule
ViT-B/16        & 52.40$\pm$4.1 & 26.63$\pm$3.2 & 34.28$\pm$2.5 & 32.95$\pm$0.9 \\
$\tau^k$        & 33.03$\pm$11.4 & 18.19$\pm$8.1 & 25.79$\pm$7.1 & 23.25$\pm$4.6 \\
Unconst.        & 19.01$\pm$12.3 & 9.89$\pm$8.5 & 13.82$\pm$11.1 & 14.32$\pm$+8.4 \\
CVX             & 34.38$\pm$14.0 & 19.94$\pm$8.6 & 24.66$\pm$10.5 & 23.23$\pm$6.12 \\
Dual (Ours)     & \textbf{71.82$\pm$4.2} & \textbf{45.23$\pm$2.0} & \textbf{48.59$\pm$2.9} & 32.75$\pm$2.3  \\
\midrule\midrule
                 & \multicolumn{4}{|c}{Block 4} \\
\midrule
DeepLIFT        & 55.66 & 40.24 & 33.38 & 20.93 \\
LRP             & 39.58 & 27.51 & 23.51 & 16.10 \\
GradCAM         & 45.04 & 23.78 & 31.87 & 29.41 \\
\midrule
$\tau^l$        & 34.17$\pm$9.0 & 17.88$\pm$4.9 & 24.20$\pm$5.2 & 23.89$\pm$4.2 \\
Unconst.        & 41.44$\pm$10.0 & 23.65$\pm$5.5 & 29.19$\pm$5.6 & 26.09$\pm$4.0 \\
$\text{KL}(\tau^l||\sum_{j \in N(i)}\tau^k)$ & 45.78$\pm$7.21 & - & - & - \\
$\text{KL}(\tau^k||\sum_{j \in N(i)}\tau^l)$ & 45.56$\pm$7.14 & - & - & - \\ 
Dual (Ours)     & \textbf{76.16$\pm$3.8} & \textbf{46.17$\pm$1.7} & \textbf{51.29$\pm$2.7} & \textbf{34.64$\pm$2.0} \\
\bottomrule
\end{tabular}
}

\end{center}
\end{table}


\section{Conclusion and future work} \label{conclusion}

Deploying machine learning algorithms in high-risk environments requires explaining their decisions to the user to ensure the model relies on the relevant anomalies for its decision, e.g., sites with abnormal FDG uptake in PET scans. Attention probability distributions are appealing as they both restrict the information flow through the neural network as well as visualize these restrictions as explanations. While setting multiple attention gates may improve the deep net's performance, it introduces inconsistencies between them. Thus, their explanatory value decreases and hinders the interpreting physician's ability to rely on the algorithm's decision. In this work, we suggest enforcing consistency using convex optimization. To avoid the computational complexity accompanying such operations, we further use a duality theorem to decompose the task into two reparameterized attention probability distributions. These reparameterizations allow us to learn consistent attention units without solving the primal convex optimization program.  

There are a few limitations to our work. First, the choice of which convolutional blocks to apply our attention gates to was not exhaustively explored, as it is exponential in the network's depth. In addition, our proposed method outputs explanations that typically have a lower resolution compared to the input. Lastly, attention mechanisms are label-agnostic by nature, i.e., it is unclear what is the explanatory value of attention weight distribution when the predicted label indicates no findings in the medical image.

This work can be extended in several directions. With the success of deep nets, explainability in high-risk environments brings forth design choices for the learning algorithms. Consistent attention is such a choice that fits medical images; however, other design choices are required for different tasks and different data modalities. Also, learning convex optimization layers provides useful inductive bias, but they are slow to adopt. Dual decomposition can be used as a leading principle to improve their adoption beyond learning consistent probability models. Such general-purpose solvers require learning complex dual witnesses and are subject to further research.

\newpage
\bibliography{main}
\bibliographystyle{icml2022}

\newpage
\appendix
\onecolumn

\end{document}